%% file: main.tex
\documentclass[10pt,twocolumn,letterpaper]{article}

\usepackage[pagenumbers]{cvpr} 

\input{preamble}

\definecolor{cvprblue}{rgb}{0.21,0.49,0.74}
\usepackage{multirow,threeparttable,algorithm,algpseudocode}
\usepackage{colortbl}  
\usepackage{xcolor}
\usepackage{array}   
\usepackage{marvosym}
\usepackage[pagebackref,breaklinks,colorlinks,citecolor=cvprblue]{hyperref}

\usepackage{caption}

\title{4D Gaussian Splatting for Real-Time Dynamic Scene Rendering}
\author{%
  Guanjun Wu$^{1}$\footnotemark[1], \quad Taoran Yi$^{2}$\footnotemark[1], \quad Jiemin Fang$^{3}$\footnotemark[2], \quad Lingxi Xie$^{3}$, \quad Xiaopeng Zhang$^{3}$,  \\
   Wei Wei$^{1}$, \quad Wenyu Liu$^{2}$, \quad  Qi Tian$^{3}$, \quad Xinggang Wang$^{2}$\footnotemark[2]\ \footnotemark[3]\\
  $^1$School of CS, Huazhong University of Science and Technology\\
  $^2$School of EIC, Huazhong University of Science and Technology \;\;
  $^3$Huawei Inc.\\
  \texttt{\small\{guajuwu, taoranyi, weiw, liuwy, xgwang\}@hust.edu.cn}\\
\texttt{\small\{jaminfong, 198808xc, zxphistory\}@gmail.com}  \;\;
\texttt{\small tian.qi1@huawei.com}
}

\begin{document}

\twocolumn[{%
\renewcommand\twocolumn[1][]{#1}%
\maketitle
\vspace{-25pt}
\begin{center}
\centering
\includegraphics[width=1\linewidth]{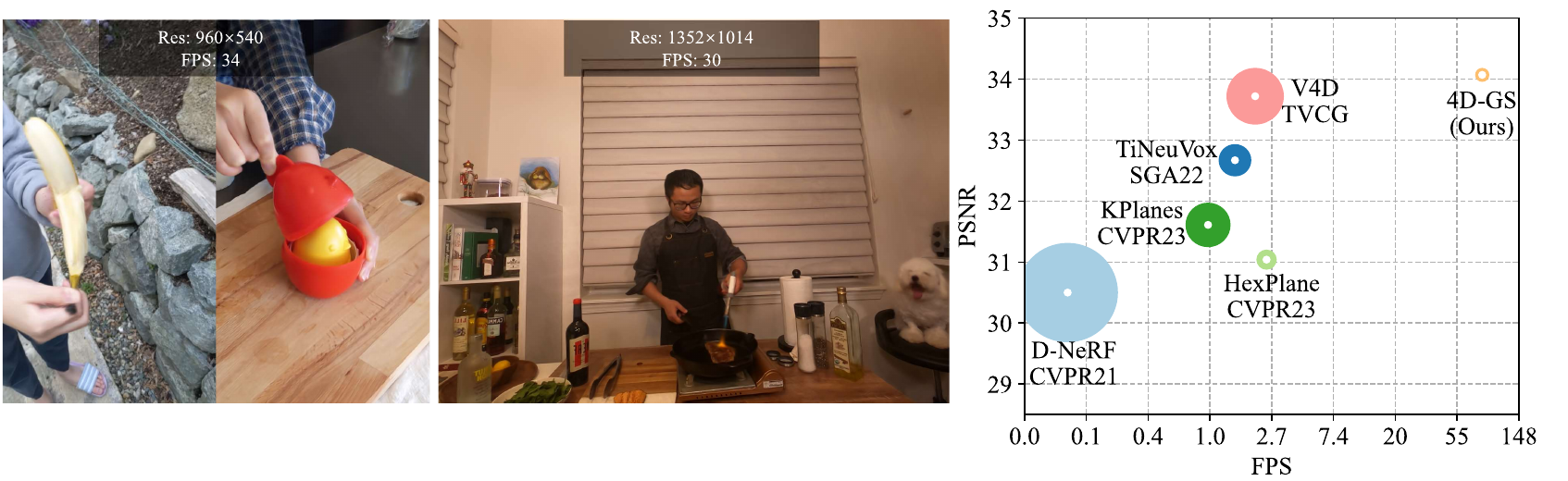}
\label{fig:teaserfig}
\vspace{-18pt}
\captionof{figure}{Our method achieves real-time rendering$^\ddag$ for dynamic scenes at high image resolutions while maintaining high rendering quality. The right figure is tested on synthetic datasets~\cite{pumarola2021dnerf}, where the radius of the dot corresponds to the training time. ``Res'': resolution.
}
\end{center}%
\vspace{-10pt}
\scriptsize $^\ddag$The rendering speed not only depends on the image resolution but also the number of 3D Gaussians and the scale of deformation fields which are determined by the complexity of the scene.
}]
\vspace{-20pt}
\footnotetext{
\llap{\textsuperscript{*}}Equal contributions.\;   \ \llap{\textsuperscript{$\dagger$}}Project lead.\;   \ \llap{\textsuperscript{$\ddagger$}}Corresponding author.
}

\begin{abstract}
\vspace{-10pt}
Representing and rendering dynamic scenes has been an important but challenging task. Especially, to accurately model complex motions, high efficiency is usually hard to guarantee. To achieve real-time dynamic scene rendering while also enjoying high training and storage efficiency, we propose 4D Gaussian Splatting (4D-GS) as a holistic representation for dynamic scenes rather than applying 3D-GS for each individual frame. In 4D-GS, a novel explicit representation containing both 3D Gaussians and 4D neural voxels is proposed. A decomposed neural voxel encoding algorithm inspired by HexPlane is proposed to efficiently build Gaussian features from 4D neural voxels and then a lightweight MLP is applied to predict Gaussian deformations at novel timestamps. Our 4D-GS method achieves real-time rendering under high resolutions, 82 FPS at an 800$\times$800 resolution on an RTX 3090 GPU while maintaining comparable or better quality than previous state-of-the-art methods. More demos and code are available at \url{https://guanjunwu.github.io/4dgs/}. 
\end{abstract}


\section{Introduction}
\label{sec:intro}
Novel view synthesis (NVS) stands as a critical task in the domain of 3D vision and plays a vital role in many applications, \eg VR, AR, and movie production. NVS aims at rendering images from any desired viewpoint or timestamp of a scene, usually requiring modeling the scene accurately from several 2D images. Dynamic scenes are quite common in real scenarios, rendering which is important but challenging as complex motions need to be modeled with both spatially and temporally sparse input.

NeRF~\cite{mildenhall2021nerf} has achieved great success in synthesizing novel view images by representing scenes with implicit functions. The volume rendering techniques~\cite{drebin1988volume} are introduced to connect 2D images and 3D scenes. However, the original NeRF method bears big training and rendering costs. Though some NeRF variants~\cite{dvgo,tineuvox,hexplane,kplanes,tensor4d,fridovich2022plenoxels,instantngp} reduce the training time from days to minutes, the rendering process still bears a non-negligible latency.

Recent 3D Gaussian Splatting (3D-GS)~\cite{3dgs} significantly boosts the rendering speed to a real-time level by representing the scene as 3D Gaussians. The cumbersome volume rendering in the original NeRF is replaced with efficient differentiable splatting~\cite{yifan2019differentiablesplatting}, which directly projects 3D Gaussian onto the 2D image plane. 3D-GS not only enjoys real-time rendering speed but also represents the scene more explicitly, making it easier to manipulate the scene representation. 

However, 3D-GS focuses on the static scenes. Extending it to dynamic scenes as a 4D representation is a reasonable, important but difficult topic. The key challenge lies in modeling complicated point motions from sparse input. 3D-GS holds a natural geometry prior by representing scenes with point-like Gaussians. One direct and effective extension approach is to construct 3D Gaussians at each timestamp~\cite{dynamic3dgs} but the storage/memory cost will multiply especially for long input sequences. Our goal is to construct a compact representation while maintaining both training and rendering efficiency, \ie 4D Gaussian Splatting (\textbf{4D-GS}). To this end, we propose to represent Gaussian motions and shape changes by an efficient Gaussian deformation field network, containing a temporal-spatial structure encoder and an extremely tiny multi-head Gaussian deformation decoder. Only one set of canonical 3D Gaussians is maintained. For each timestamp, the canonical 3D Gaussians will be transformed by the Gaussian deformation field into new positions with new shapes. The transformation process represents both the Gaussian motion and deformation. Note that different from modeling motions of each Gaussian separately~\cite{dynamic3dgs,4dgs-2}, the spatial-temporal structure encoder can connect different adjacent 3D Gaussians to predict more accurate motions and shape deformation. Then the deformed 3D Gaussians can be directly splatted for rendering the according-timestamp image. Our contributions can be summarized as follows.

\begin{itemize}
     \item An efficient 4D Gaussian splatting framework with an efficient Gaussian deformation field is proposed by modeling both Gaussian motion and Gaussian shape changes across time.
     \item A multi-resolution encoding method is proposed to connect the nearby 3D Gaussians and build rich 3D Gaussian features by an efficient spatial-temporal structure encoder. 
     \item 4D-GS achieves real-time rendering on dynamic scenes, up to 82 FPS at a resolution of 800$\times$800 for synthetic datasets and 30 FPS at a resolution of 1352$\times$1014 in real datasets, while maintaining comparable or superior performance than previous state-of-the-art (SOTA) methods. It also shows potential for editing and tracking in 4D scenes.
\end{itemize}

\begin{figure}
   \centering
   \includegraphics[width=1.0\linewidth]{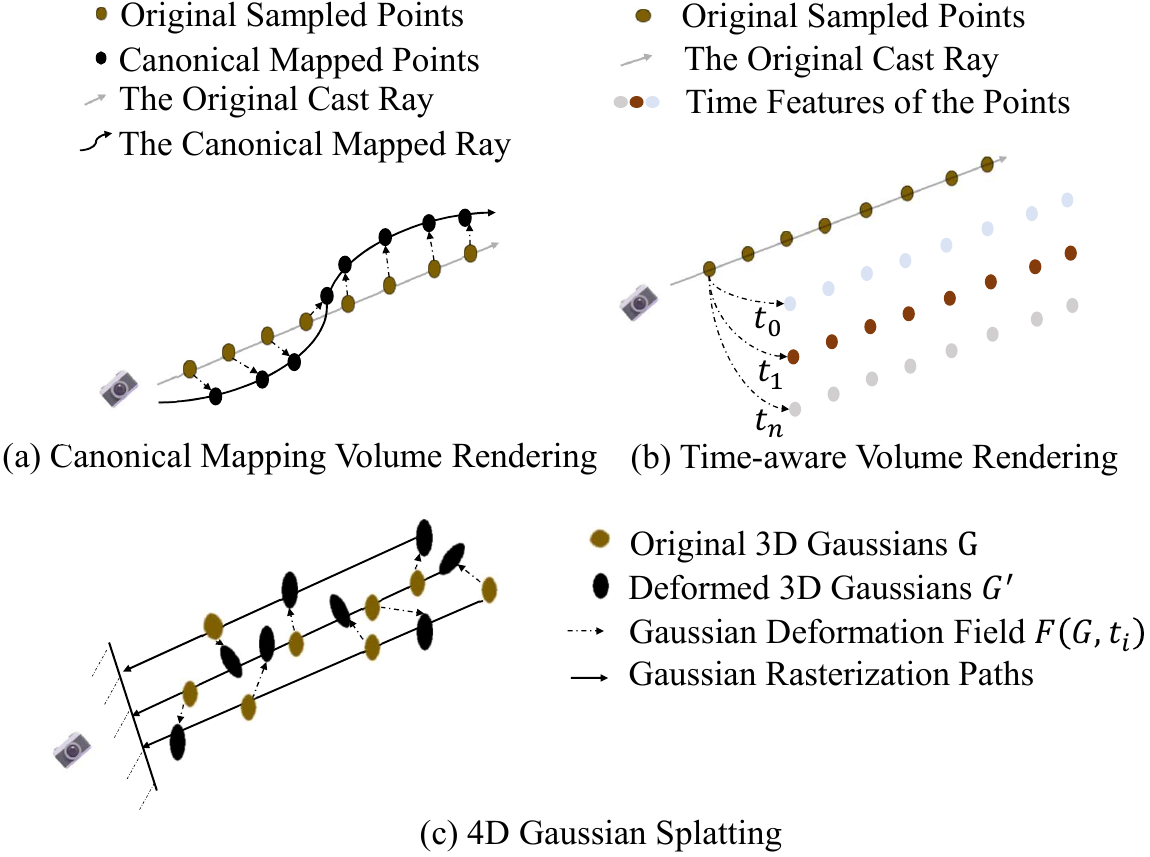}
   \caption{Illustration of different dynamic scene rendering methods. (a) Points are sampled in the cast ray during volume rendering. The point deformation fields proposed in~\cite{pumarola2021dnerf, tineuvox} map the points into a canonical space. (b) Time-aware volume rendering computes the features of each point directly and does not change the rendering path. (c) The Gaussian deformation field converts original 3D Gaussians into another group of 3D Gaussians with a certain timestamp.}
   \label{fig: splatting-volume}
   \vspace{-10pt}
\end{figure}

\begin{figure*}[thbp]
   \centering
   \includegraphics[width=1.0\linewidth]{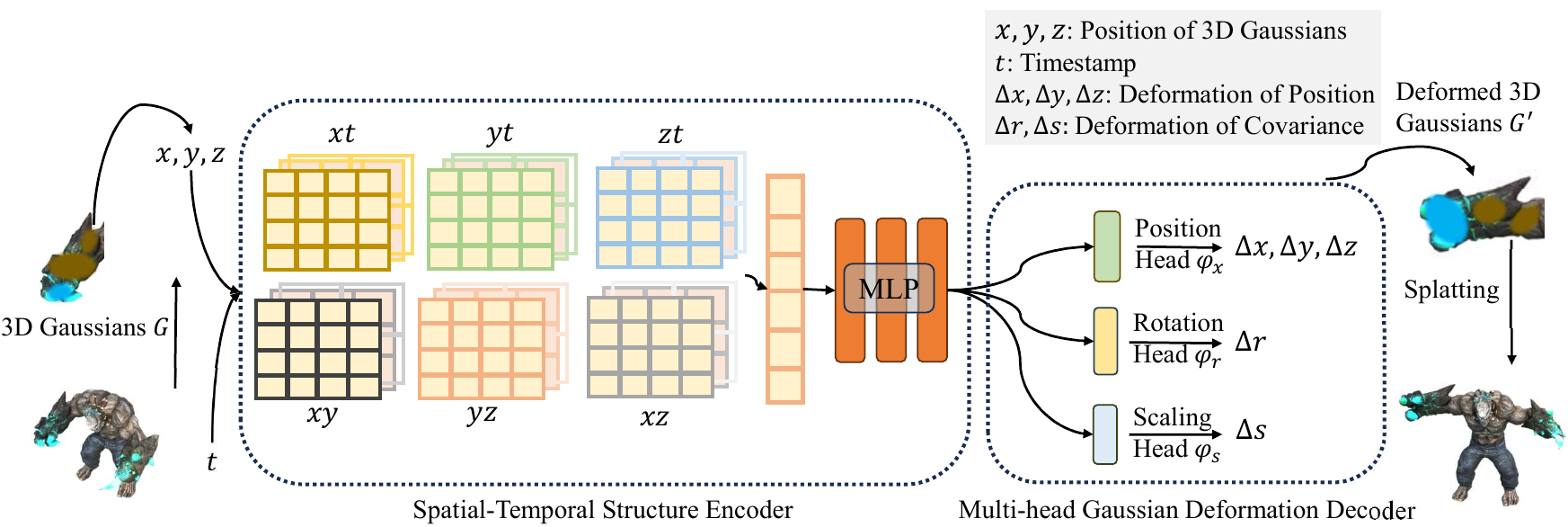}
   \caption{The overall pipeline of our model. Given a group of 3D Gaussians $\mathcal{G}$, we extract the center coordinate of each 3D Gaussian $\mathcal{X}$ and timestamp $t$ to compute the voxel feature by querying multi-resolution voxel planes. Then a tiny multi-head Gaussian deformation decoder is used to decode the feature and get the deformed 3D Gaussians $\mathcal{G}^\prime$ at timestamp $t$. The deformed Gaussians are then splatted to get the rendered images.}
   \label{fig: pipeline}
   \vspace{-10pt}
\end{figure*}
\section{Related Works}
In this section, we simply review the difference of dynamic NeRFs in Sec.~\ref{subsec:dynamic novel view synthesis}, then discuss the point clouds-based neural rendering algorithm in Sec.~\ref{subsec:neural rendering with point clouds}.

\subsection{Novel View Synthesis}
\label{subsec:dynamic novel view synthesis}
Novel view synthesis is a important and challenging task in 3D reconstruction. Much approaches are proposed to represent a 3D object and render novel views. Efficient representations such as light fields~\cite{broxton2020immersive}, mesh~\cite{collet2015high,li20184d,guo2015robust,su2020robustfusion}, voxels~\cite{guo2019relightables,hu2022hvtr,li2017robust}, multi-planes~\cite{flynn2019deepview} can render high quality image with enough supervisions. NeRF-based approaches~\cite{mildenhall2021nerf,barron2021mip,zhang2020nerf++} demonstrate that implicit radiance fields can effectively learn scene representations and synthesize high-quality novel views.~\cite{pumarola2021dnerf,park2021nerfies,park2021hypernerf} have challenged the static hypothesis, expanding the boundary of novel view synthesis for dynamic scenes.~\cite{tineuvox} proposes to use an explicit voxel grid to model temporal information, accelerating the learning time for dynamic scenes to half an hour and applied in~\cite{gneuvox,robustdynamicradiancefields,guo2023forwardflowfornvs}. The proposed deformation-based neural rendering methods are shown in Fig.~\ref{fig: splatting-volume} \textcolor{red} {(a)}. Flow-based~\cite{gao2021dynamic,tian2023mononerf,li2021neural,robustdynamicradiancefields,zhou2024dynpoint} methods adopting warping algorithm to synthesis novel views by blending nearby frames. ~\cite{li2022neural,hexplane,kplanes,tensor4d,msth,gan2023v4d} represent further advancements in faster dynamic scene learning by adopting decomposed neural voxels. They treat sampled points in each timestamp individually as shown in Fig.~\ref{fig: splatting-volume} \textcolor{red} {(b)}.~\cite{wang2023neus2,gao2022mps,mlpmaps,xu2022multigeometric,lin2023im4d,wang2023mixedvoxels} are efficient methods to handle multi-view setups. The aforementioned methods though achieve fast training speed, real-time rendering for dynamic scenes is still challenging, especially for monocular input. Our method aims at constructing a highly efficient training and rendering pipeline in Fig.~\ref{fig: splatting-volume} \textcolor{red} {(c)}, while maintaining the quality, even for sparse inputs.


\subsection{Neural Rendering with Point Clouds}\label{subsec:neural rendering with point clouds}
Effectively representing 3D scenes remains a challenging topic. The community has explored various neural representations~\cite{mildenhall2021nerf}, \eg meshes, point clouds~\cite{pointnerf}, voxels~\cite{fridovich2022plenoxels}, and hybrid approaches~\cite{dvgo,instantngp}. Point-cloud-based methods~\cite{qi2017pointnet,qi2017pointnet++,yu2018pu,liu2019meteornet} initially target at 3D segmentation and classification. A representative approach for rendering presented in~\cite{pointnerf, abou2022particlenerf} combines point cloud representations with volume rendering, achieving rapid convergence speed even for dynamic novel view synthesis~\cite{zhou2024dynpoint,park2024point}. ~\cite{keselman2023flexibleTechniquesDifferentiableRendering,keselman2022approximatedifferentiablerenderingwithalgebraicsurfaces,ruckert2022adop} adopt differential point rendering technique for scene reconstructions. 

Recently, 3D-GS~\cite{3dgs,chen2024survey} is notable for its pure explicit representation and differential point-based splatting methods, enabling real-time rendering of novel views. Dynamic3DGS~\cite{dynamic3dgs} models dynamic scenes by tracking the position and variance of each 3D Gaussian at each timestamp $t_i$. An explicit table is utilized to store information about each 3D Gaussian at every timestamp, leading to a linear memory consumption increase, denoted as $O(t\mathcal{N})$, in which $\mathcal{N}$ is num of 3D Gaussians. For long-term scene reconstruction, the storage cost will become non-negligible. The memory complexity of our approach only depends on the number of 3D Gaussians and parameters of Gaussians deformation fields network $\mathcal{F}$, which is denoted as $O(\mathcal{N}+\mathcal{F})$. Another method to extend 3D Gaussians to 4D~\cite{4dgs-2} adds a marginal temporal Gaussian distribution into the origin 3D Gaussians, which uplifts 3D Gaussians into 4D. However, it may cause each 3D Gaussian to only focus on their local temporal space. Deformable-3DGS~\cite{yang2023deformable3dgs} is a concurrent work that introduces an MLP deformation network to model the motion of dynamic scenes. Spacetime-GS~\cite{li2023spacetime} tracks each 3D Gaussians individually. Our approach also models 3D Gaussian motions but with a compact network, resulting in highly efficient training and real-time rendering.  

\section{Preliminary}
In this section, we simply review the representation and rendering process of 3D-GS~\cite{3dgs} in Sec.~\ref{subsec:3D Gaussian Splatting} and the formula of dynamic NeRFs in Sec.~\ref{subsec:Dynamic NeRFs with deformation Fields.}.

\subsection{3D Gaussian Splatting}
\label{subsec:3D Gaussian Splatting}
3D Gaussians~\cite{3dgs} is an explicit 3D scene representation in the form of point clouds. Each 3D Gaussian is characterized by a covariance matrix $\Sigma$ and a center point $\mathcal{X}$, which is referred to as the mean value of the Gaussian:
\begin{equation}
\label{formula:gaussian's formula}
    G(X)=e^{-\frac{1}{2}\mathcal{X}^T\Sigma^{-1}\mathcal{X}}.
\end{equation}
For differentiable optimization, the covariance matrix $\Sigma$ can be decomposed into a scaling matrix $\mathbf{S}$ and a rotation matrix $\mathbf{R}$:
\begin{equation}
\label{formula:covariance decomposition}
    \Sigma = \mathbf{R}\mathbf{S}\mathbf{S}^T\mathbf{R}^T.
\end{equation}

When rendering novel views, differential splatting~\cite{yifan2019differentiablesplatting} is employed for the 3D Gaussians within the camera planes. As introduced by~\cite{zwicker2001surfacesplatting}, using a viewing transform matrix $W$ and the Jacobian matrix $J$ of the affine approximation of the projective transformation, the covariance matrix $\Sigma^{\prime}$ in camera coordinates can be computed as
\begin{equation}
    \Sigma^{\prime} = JW\Sigma W^TJ^T.
\end{equation}
In summary, each 3D Gaussian is characterized by the following attributes: position $\mathcal{X} \in \mathbb{R}^3$, color defined by spherical harmonic (SH) coefficients $\mathcal{C} \in \mathbb{R}^k$ (where $k$ represents nums of SH functions), opacity $\alpha \in \mathbb{R}$, rotation factor $r \in \mathbb{R}^4$, and scaling factor $s \in \mathbb{R}^3$.
Specifically, for each pixel, the color and opacity of all the Gaussians are computed using the Gaussian's representation Eq.~\ref{formula:gaussian's formula}. The blending of $N$ ordered points that overlap the pixel is given by the formula:
\begin{equation}
\label{formula: splatting&volume rendering}
    C = \sum_{i\in N}c_i \alpha_i \prod_{j=1}^{i-1} (1-\alpha_i).
\end{equation}
Here, $c_i$, $\alpha_i$ represents the density and color of this point computed by a 3D Gaussian $G$ with covariance $\Sigma$ multiplied by an optimizable per-point opacity and SH color coefficients.

\subsection{Dynamic NeRFs with Deformation Fields}
\label{subsec:Dynamic NeRFs with deformation Fields.}
All the dynamic NeRF algorithms can be formulated as:
\begin{equation}
    c,\sigma = \mathcal{M}(\mathbf{x},d,t,\lambda),
\end{equation}
where $\mathcal{M}$ is a mapping that maps 8D space $(\mathbf{x},d,t,\lambda)$ to 4D space $(c,\sigma)$. $\mathbf{x}$ reveals to the spatial point, and $\lambda$ is the optional input as used to build topological and appearance changes in~\cite{park2021hypernerf}, and $d$ stands for view-dependency.

As shown in Fig.~\ref{fig: splatting-volume}~\textcolor{red}{(a)}, all the deformation NeRF based methods estimate the \textbf{world-to-canonical mapping} by a deformation network $\phi_t:(\mathbf{x},t)\rightarrow \Delta \mathbf{x}$. Then a network is introduced to compute volume density and view-dependent RGB color from each ray. The formula for rendering can be expressed as:
\begin{equation}
    c, \sigma = \text{NeRF}(\mathbf{x}+\Delta \mathbf{x},d,\lambda),
\end{equation}
where `NeRF' stands for vanilla NeRF pipeline, $\lambda$ is a frame-dependent code to model the topological and appearance changes~\cite{park2021hypernerf,martin2021nerf}.

However, our 4D Gaussian splatting framework presents a novel rendering technique. We successfully compute the \textbf{canonical-to-world mapping} directly at time $t$ using a Gaussian deformation field network $\mathcal{F}$, and differential splatting~\cite{3dgs} follows. This enables the capability of computing backward flow and tracking for 3D Gaussians.



\section{Method}
\begin{figure}
   \centering
   \includegraphics[width=\linewidth]{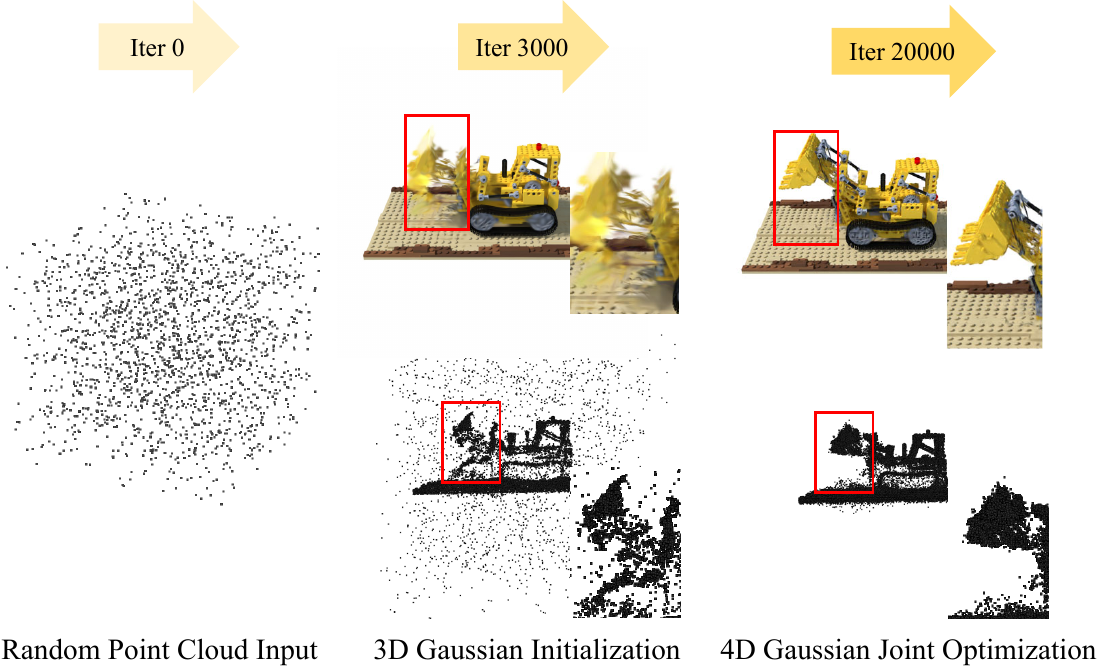}
   \caption{Illustration of the optimization process. With static 3D Gaussian initialization, our model can learn high-quality 3D Gaussians of the motion part. }
   \label{fig: coarse2fine}
\end{figure}
Sec.~\ref{subsec:4dgs} introduces the overall 4D Gaussian Splatting framework. Then, the Gaussian deformation field is proposed in Sec.~\ref{subsec:filed}. Finally, we describe the optimization process in Sec.~\ref{subsec:optimization}. 

\subsection{4D Gaussian Splatting Framework}
\label{subsec:4dgs}

As shown in Fig.~\ref{fig: pipeline}, given a view matrix $M=[R, T]$, timestamp $t$, our 4D Gaussian splatting framework includes 3D Gaussians $\mathcal{G}$ and Gaussian deformation field network $\mathcal{F}$. Then a novel-view image $\hat{I}$ is rendered by differential splatting~\cite{yifan2019differentiablesplatting} $\mathcal{S}$ following $\hat{I} = \mathcal{S}(M,\mathcal{G}^\prime)$, where $\mathcal{G}^\prime=\Delta\mathcal{G}+\mathcal{G}$.

Specifically, the deformation of 3D Gaussians $\Delta \mathcal{G}$ is introduced by the Gaussian deformation field network $\Delta \mathcal{G} = \mathcal{F}(\mathcal{G}, t)$, in which the spatial-temporal structure encoder $\mathcal{H}$ can encode both the temporal and spatial features of 3D Gaussians $f_d=\mathcal{H}(\mathcal{G}, t)$. And the multi-head Gaussian deformation decoder $\mathcal{D}$ can decode the features and predict each 3D Gaussian's deformation $\Delta \mathcal{G} = \mathcal{D}(f)$, then the deformed 3D Gaussians $\mathcal{G}^\prime$ can be introduced.


The rendering process of our 4D Gaussian Splatting is depicted in Fig.~\ref{fig: splatting-volume}~\textcolor{red}{(c)}. Our 4D Gaussian splatting converts the original 3D Gaussians $\mathcal{G}$ into another group of 3D Gaussians $\mathcal{G}^\prime$ given a timestamp $t$, maintaining the effectiveness of the differential splatting as referred in~\cite{yifan2019differentiablesplatting}.


\begin{figure*}
   \centering
   \includegraphics[width=1.0\linewidth]{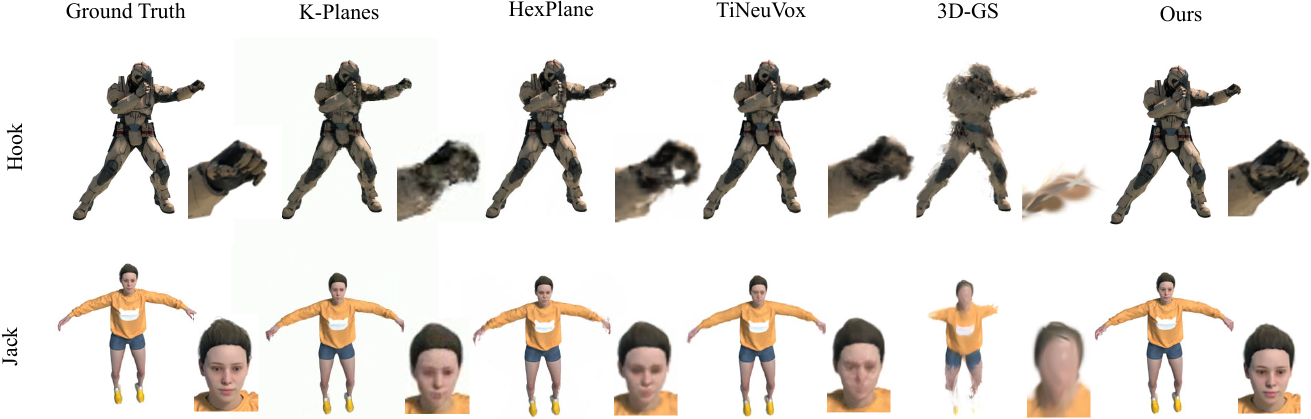}
   \caption{Visualization of synthesized datasets compared with other models~\cite{hexplane,kplanes,tineuvox,3dgs,guo2023forwardflowfornvs,msth}. The rendering results of~\cite{kplanes} are displayed with a default green background. We adopt their rendering settings.}
      \label{fig: dnerf_result}
      \vspace{-10pt}
\end{figure*}

\subsection{Gaussian Deformation Field Network}
\label{subsec:filed}

The network to learn the Gaussian deformation field includes an efficient spatial-temporal structure encoder $\mathcal{H}$ and a Gaussian deformation decoder $\mathcal{D}$ for predicting the deformation of each 3D Gaussian.

\paragraph{Spatial-Temporal Structure Encoder.}

Nearby 3D Gaussians always share similar spatial and temporal information. To model 3D Gaussians' features effectively, we introduce an efficient spatial-temporal structure encoder $\mathcal{H}$ including a multi-resolution HexPlane $R(i,j)$ and a tiny MLP $\phi_d$ inspired by ~\cite{tineuvox,hexplane,kplanes,tensor4d}. While the vanilla 4D neural voxel is memory-consuming, we adopt a 4D K-Planes~\cite{kplanes} module to decompose the 4D neural voxel into 6 multi-resolution planes. All 3D Gaussians in a certain area can be contained in the bounding plane voxels and the deformation of Gaussians can also be encoded in nearby temporal voxels.

Specifically, the spatial-temporal structure encoder $\mathcal{H}$ contains 6 multi-resolution plane modules $R_l(i,j)$ and a tiny MLP $\phi_d$, \ie $\mathcal{H}(\mathcal{G},t)=\{R_l(i,j), \phi_d | (i,j) \in \{(x,y),(x,z),(y,z),(x,t),(y,t),(z,t)\}, l \in \{1,2\}\}$. The position $\mu = (x,y,z)$ is the mean value of 3D Gaussians $\mathcal{G}$. Each voxel module is defined by $R(i,j)\in \mathbb{R}^{h\times l N_i \times l N_j}$, where $h$ stands for the hidden dim of features, and $N$ denotes the basic resolution of voxel grid and $l$ equals to the upsampling scale. 
This entails encoding information of the 3D Gaussians within the 6 2D voxel planes while considering temporal information. The formula for computing separate voxel features is as follows:
\begin{equation}
\begin{aligned}
  f_h &= \bigcup_l \prod \text{interp}(R_l(i,j)), \\
    (i,j) &\in \{(x,y),(x,z),(y,z),(x,t),(y,t),(z,t)\}. 
\end{aligned}
\label{formula:feature_position}
\end{equation}
$f_{h} \in \mathbb{R}^{h*l}$ is the feature of neural voxels. `interp' denotes the bilinear interpolation for querying the voxel features located at 4 vertices of the grid. The discussion of the production process is similar to K-Planes~\cite{kplanes}. Then a tiny MLP $\phi_d$ merges all the features by $f_d = \phi_d(f_h)$.


\paragraph{Multi-head Gaussian Deformation Decoder.}
\label{subsec:Gaussian Deformation Decoder}

When all the features of 3D Gaussians are encoded, we can compute any desired variable with a multi-head Gaussian deformation decoder $\mathcal{D}=\{\phi_x, \phi_r, \phi_s\}$. Separate MLPs are employed to compute the deformation of position $\Delta \mathcal{X} = \phi_x(f_d)$, rotation $\Delta r= \phi_r(f_d)$, and scaling $\Delta s=\phi_s(f_d)$.
Then, the deformed feature $(\mathcal{X}^\prime,r^\prime,s^\prime)$ can be addressed as:
\begin{align}
    & (\mathcal{X}^\prime, r^\prime, s^\prime) = (\mathcal{X} + \Delta \mathcal{X}, r + \Delta r, s + \Delta s) .
\end{align}
Finally, we obtain the deformed 3D Gaussians $\mathcal{G}^\prime=\{\mathcal{X}^\prime, s^\prime, r^\prime,\sigma,\mathcal{C} \}$.



\subsection{Optimization}
\label{subsec:optimization}

\paragraph{3D Gaussian Initialization.}
3D-GS~\cite{3dgs} shows that 3D Gaussians can be well-trained with structure from motion (SfM)~\cite{schonberger2016structure} points initialization. Similarly, 4D Gaussians can also leverage the power of proper 3D Gaussian initialization. We optimize 3D Gaussians at initial 3000 iterations for warm-up and then render images with 3D Gaussians $\hat{I} = \mathcal{S}(M,\mathcal{G})$ instead of 4D Gaussians $\hat{I}=\mathcal{S}(M,\mathcal{G}^\prime)$. The illustration of the optimization process is shown in Fig.~\ref{fig: coarse2fine}.

\paragraph{Loss Function.}
Similar to other reconstruction methods~\cite{3dgs,pumarola2021dnerf,tineuvox}, we use the L1 color loss to supervise the training process. A grid-based total-variational loss~\cite{dvgo,tineuvox,hexplane,kplanes} $\mathcal{L}_{tv}$ is also applied.
\begin{equation}
    \mathcal{L} = \lvert \hat{I} - I \vert + \mathcal{L}_{tv}.
\end{equation}

\begin{figure*}
   \centering
   \includegraphics[width=\linewidth]{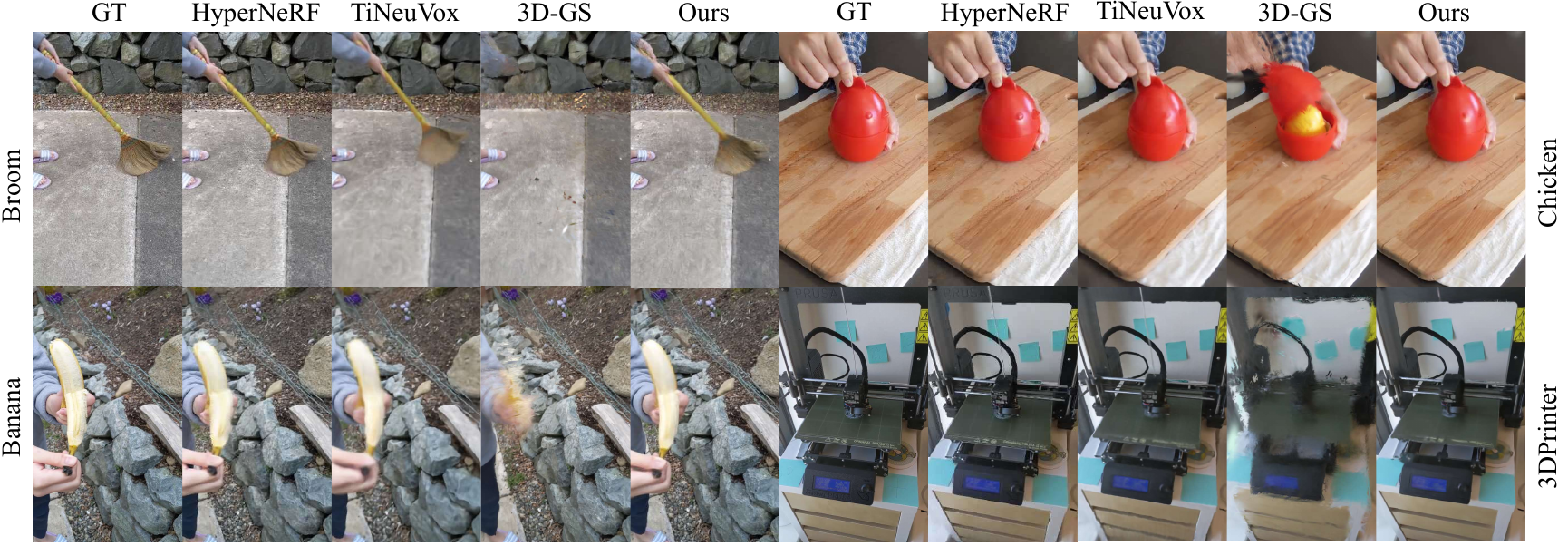}
    \caption{Visualization of the HyperNeRF~\cite{park2021hypernerf} dataset compared with other methods~\cite{3dgs,park2021hypernerf,tineuvox,guo2023forwardflowfornvs}. `GT' stands for ground truth images.}
      \label{fig:hyper_result}
\end{figure*}


\vspace{-10pt}
\begin{table*} 
\centering

\caption{Quantitative results on the synthetic dataset. The \colorbox{pink}{best} and the \colorbox{yellow}{second best} results are denoted by pink and yellow. The rendering resolution is set to 800$\times$800. ``Time'' in the table stands for training times.}
\scalebox{0.9}{
\begin{tabular}{l|ccc|c|cc} 
\toprule
Model  & PSNR (dB)↑ & SSIM↑ & LPIPS↓ & Time↓ &  FPS ↑ & Storage (MB)↓  \\
\midrule  
TiNeuVox-B~\cite{tineuvox}  & 32.67 & 0.97 &  0.04 & 28 mins & 1.5 & 48\\ 
KPlanes~\cite{kplanes} & 31.61 & 0.97 &-& 52 mins& 0.97   &418\\ 
HexPlane-Slim~\cite{hexplane} & 31.04 & 0.97 & 0.04&11m 30s& 2.5 &38\\ 
3D-GS~\cite{3dgs} &23.19 & 0.93 & 0.08&10 mins  &\cellcolor{pink}170&\cellcolor{pink}10 \\
FFDNeRF~\cite{guo2023forwardflowfornvs}&32.68&0.97&0.04&-&$<$ 1&440\\
MSTH~\cite{msth}&31.34&0.98&0.02&\cellcolor{pink}6 mins&-&-\\
V4D~\cite{gan2023v4d}&33.72\cellcolor{yellow}&0.98\cellcolor{yellow}&0.02\cellcolor{yellow}&6.9 hours &2.08&377\\
Ours & \cellcolor{pink}34.05&\cellcolor{pink}0.98 &\cellcolor{pink}0.02\cellcolor{pink}& 8 mins\cellcolor{yellow}&\cellcolor{yellow}82 &\cellcolor{yellow}18 \\
\bottomrule

\end{tabular}  
}

\label{table_dnerf}
\end{table*}
\vspace{-5pt}

\begin{table*} 
\centering
\small
\caption{Quantitative results on HyperNeRF~\cite{park2021hypernerf} vrig dataset with the rendering resolution of 960$\times$540.}
\scalebox{1.0}{
\begin{tabular}{l|cc|c|cc} 
\toprule  
Model  & PSNR (dB)↑ & MS-SSIM↑  & Times↓ &  FPS↑ & Storage (MB)↓\\ 
\midrule  
Nerfies~\cite{park2021nerfies}&22.2&0.803&$\sim$ hours&$<$ 1&-\\
HyperNeRF~\cite{park2021hypernerf}&22.4&0.814&32 hours &$<$ 1&-\\
TiNeuVox-B~\cite{tineuvox}&24.3&0.836&\cellcolor{yellow}30 mins&1&\cellcolor{pink}48\\
3D-GS~\cite{3dgs}&19.7&0.680&40 mins&\cellcolor{pink}55&\cellcolor{yellow}52\\
FFDNeRF~\cite{guo2023forwardflowfornvs}&24.2&\cellcolor{yellow}0.842&-&0.05&440\\
V4D~\cite{gan2023v4d}&24.8\cellcolor{yellow}&0.832&5.5 hours&0.29&377\\
Ours&\cellcolor{pink}25.2&\cellcolor{pink}0.845&\cellcolor{pink}30 mins& \cellcolor{yellow}34&61\\
\bottomrule
\end{tabular}  
}
\label{tab:hypernerf}
\end{table*}

\section{Experiment}

In this section, we mainly introduce the hyperparameters and datasets of our settings in Sec.~\ref{Experimental Setting} and the results between different datasets are compared with \cite{tineuvox,msth,wang2023mixedvoxels,lin2023im4d,attal2023hyperreel,hexplane,kplanes,3dgs,song2023nerfplayer} in Sec.~\ref{subsec:result}. Then, ablation studies are proposed to prove the effectiveness of our approach in Sec.~\ref{subsec:ablation study} and more discussion about 4D-GS in Sec.~\ref{subsec:discussions}. Finally, we discuss the limitation of our proposed 4D-GS in Sec.~\ref{Subsec:limitation}.
\subsection{Experimental Settings}
\label{Experimental Setting}
Our implementation is primarily based on the PyTorch~\cite{pytorch} framework and tested on a single RTX 3090 GPU, and we've fine-tuned our optimization parameters by the configuration outlined in the 3D-GS~\cite{3dgs}. More hyperparameters are shown in the appendix.


\paragraph{Synthetic Dataset.}We primarily assess the performance of our model using a synthetic dataset, as introduced by D-NeRF~\cite{pumarola2021dnerf}. The dataset is designed for monocular settings, although it's worth noting that the camera poses for each timestamp are close to randomly generated. Each scene within these datasets contains dynamic frames, ranging from 50 to 200 in number.
\paragraph{Real-world Datasets.} We utilize datasets provided by HyperNeRF~\cite{park2021hypernerf} and Neu3D~\cite{li2022neural} as benchmark datasets to evaluate the performance of our model in real-world scenarios. The HyperNeRF~\cite{park2021hypernerf} dataset is captured using one or two cameras, following straightforward camera motion, while the Neu3D dataset is captured using 15 to 20 static cameras, involving extended periods and intricate camera motions. We use the points computed by SfM~\cite{schonberger2016structure} from the first frame of each video in the Neu3D dataset and 200 frames randomly selected in the HyperNeRF dataset.
\subsection{Results}
\label{subsec:result}
We primarily assess our experimental results using various metrics, encompassing peak-signal-to-noise ratio (PSNR), perceptual quality measure LPIPS~\cite{lpips}, structural similarity index (SSIM)~\cite{ssim} and its extensions including structural dissimilarity index measure (DSSIM), multiscale structural similarity index (MS-SSIM), FPS, training times and storage. 
\begin{table*} 
\centering
\caption{Quantitative results on the Neu3D~\cite{li2022neural} dataset with the rendering resolution of 1352$\times$1014.}
\scalebox{0.9}{
\setlength{\tabcolsep}{8pt}
\begin{tabular}{l|ccc|c|cc}  
\toprule
Model  & PSNR (dB)↑ & D-SSIM↓ & LPIPS↓ & Time ↓  & FPS↑ & Storage (MB)↓ \\ 
\midrule 
NeRFPlayer~\cite{song2023nerfplayer}  & 30.69 &  0.034  & 0.111  &  6 hours& 0.045&-\\  
HyperReel~\cite{attal2023hyperreel} &31.10&0.036&0.096&9 hours&2.0&360\\
HexPlane-all*~\cite{hexplane} & 31.70 & \cellcolor{pink}0.014 & 0.075 & 12 hours &0.2&250\\ 
KPlanes~\cite{kplanes}& 31.63& -   &  - &   1.8 hours  &0.3&309\\
Im4D~\cite{lin2023im4d}&\cellcolor{pink}32.58&-&0.208&\cellcolor{yellow}28 mins&$\sim$5&\cellcolor{yellow}93\\
MSTH~\cite{msth}&\cellcolor{yellow}32.37&\cellcolor{yellow}0.015&\cellcolor{yellow}0.056&\cellcolor{pink}20 mins & \cellcolor{yellow}2 (15$^\ddag$)&135\\
Ours& 31.15 & 0.016 &\cellcolor{pink}0.049 & 40 mins  &\cellcolor{pink}30 & \cellcolor{pink}90\\
\hline  
\end{tabular}  
}
 \begin{tablenotes}
        \footnotesize
        \item[*] *: The metrics of the models are tested without ``coffee martini'' and resolution is set to 1024$\times$768. 
        \item [$^\ddag$] $^\ddag$: The FPS is tested with fixed-view rendering.
        \end{tablenotes}\label{tab:dynerf}

\end{table*}
To assess the quality of novel view synthesis, we conduct comparisons with several state-of-the-art methods in the field, including ~\cite{tineuvox,hexplane,kplanes,msth,3dgs,gan2023v4d,guo2023forwardflowfornvs,lin2023im4d,park2021nerfies,park2021hypernerf,song2023nerfplayer,attal2023hyperreel}.
The K-Planes results on the synthetic dataset originate from the Deformable-3DGS~\cite{yang2023deformable3dgs} paper. The other results of the compared methods are from their papers, reproduced by their code or provided by the authors. 
The rendering speed and storage data for \cite{kplanes, hexplane, tineuvox, 3dgs} are estimated based on the official implementations.

The results in synthetic dataset~\cite{pumarola2021dnerf} are summarized in Tab.~\ref{table_dnerf}. While current dynamic hybrid representations can produce high-quality results, they often come with the drawback of rendering speed. The lack of modeling dynamic motion part makes 3D-GS~\cite{3dgs} fail to reconstruct dynamic scenes. In contrast, our method enjoys both the highest rendering quality within the synthetic dataset and exceptionally fast rendering speeds while keeping extremely low storage consumption and convergence time.

Additionally, the results obtained from real-world datasets are presented in Tab.~\ref{tab:hypernerf} and Tab.~\ref{tab:dynerf}.  It becomes apparent that some NeRFs~\cite{song2023nerfplayer,attal2023hyperreel,hexplane} suffer from slow convergence speed, and the other grid-based NeRF methods~\cite{tineuvox,hexplane,msth,kplanes} encounter difficulties when attempting to capture intricate object details. In stark contrast, our methods research comparable rendering quality, fast convergence, and excel in free-view rendering speed in indoor cases. Though Im4D~\cite{lin2023im4d} addresses the high quality in comparison to ours, the need for multi-cam setups makes it hard to model monocular scenes and other methods~\cite{msth,kplanes,hexplane,attal2023hyperreel,song2023nerfplayer} also limit free-view rendering speed and storage.
\vspace{-5pt}
\subsection{Ablation Study}
\label{subsec:ablation study}
\paragraph{Spatial-Temporal Structure Encoder.}
The explicit HexPlane encoder $R_l(i,j)$ possesses the capacity to retain 3D Gaussians' spatial and temporal information, which can reduce storage consumption in comparison with purely explicit method~\cite{dynamic3dgs}. Discarding this module, we observe that using only a shallow MLP $\phi_d$ falls short in modeling complex deformations across various settings. Tab.~\ref{table_ab} demonstrates that, while the model incurs minimal memory costs, it does come at the expense of rendering quality. 


\vspace{-5pt}
\paragraph{Gaussian Deformation Decoder.}
Our proposed Gaussian deformation decoder $\mathcal{D}$ decodes the features from the spatial-temporal structure encoder $\mathcal{H}$. All the changes in 3D Gaussians can be explained by separate MLPs $\{\phi_x,\phi_r,\phi_s\}$. As shown in Tab.~\ref{table_ab}, 4D Gaussians cannot fit dynamic scenes well without modeling 3D Gaussian motion. Meanwhile, the movement of human body joints is typically manifested as stretching and twisting of surface details in a macroscopic view. If one aims to accurately model these movements, the size and shape of 3D Gaussians should also be adjusted accordingly. Otherwise, there may be underfitting of details during excessive stretching, or an inability to correctly simulate the movement of objects at a microscopic level. 
\vspace{-5pt}
\paragraph{3D Gaussian Initialization.}
In some cases without SfM~\cite{schonberger2016structure} points initialization, training 4D-GS directly may cause difficulty in convergence. Optimizing 3D Gaussians for warm-up enjoys: (a) making some 3D Gaussians stay in the dynamic part, which releases the pressure of large deformation learning by 4D Gaussians as shown in Fig.~\ref{fig: coarse2fine}. (b) learning proper 3D Gaussians $\mathcal{G}$ and suggesting deformation fields paying more attention to the dynamic part. 
(c) avoiding numeric errors in optimizing the Gaussian deformation network $\mathcal{F}$ and keeping the training process stable. Tab.~\ref{table_ab} also shows that if we train our model without the warm-up coarse stage, the rendering quality will suffer.
\begin{table*} 
\centering
\caption{Ablation studies on synthetic datasets using our proposed methods.}
\begin{tabular}{l|ccc|c|ccc} 
\toprule
Model  & PSNR(dB)↑ & SSIM↑ & LPIPS↓ & Time↓ &FPS↑& Storage (MB)↓&\\
\midrule 
Ours w/o HexPlane $R_l(i,j)$ & 27.05 & 0.95 & 0.05& \cellcolor{pink}4 mins &\cellcolor{pink}140 &\cellcolor{pink}12 \\ 
Ours w/o initialization  &31.91 & 0.97 & 0.03&\cellcolor{yellow}7.5 mins& 79&18 \\ 
Ours w/o $\phi_x$ &26.67&0.95&0.07&8 mins&82&17\\
Ours w/o $\phi_r$ &33.08\cellcolor{yellow}&\cellcolor{yellow}0.98&\cellcolor{yellow}0.03&8 mins&83\cellcolor{yellow}&\cellcolor{yellow}17\\
Ours w/o $\phi_s$ &33.02&0.98&0.03&8 mins&82&17\\ 
Ours & \cellcolor{pink}34.05&\cellcolor{pink}0.98 &\cellcolor{pink}0.02&8 mins&82 &18 \\

\bottomrule
\end{tabular}  
\label{table_ab}
\vspace{-10pt}
\end{table*}
\subsection{Discussions}\label{subsec:discussions}
\begin{figure}
   \centering
   \includegraphics[width=0.9\linewidth]{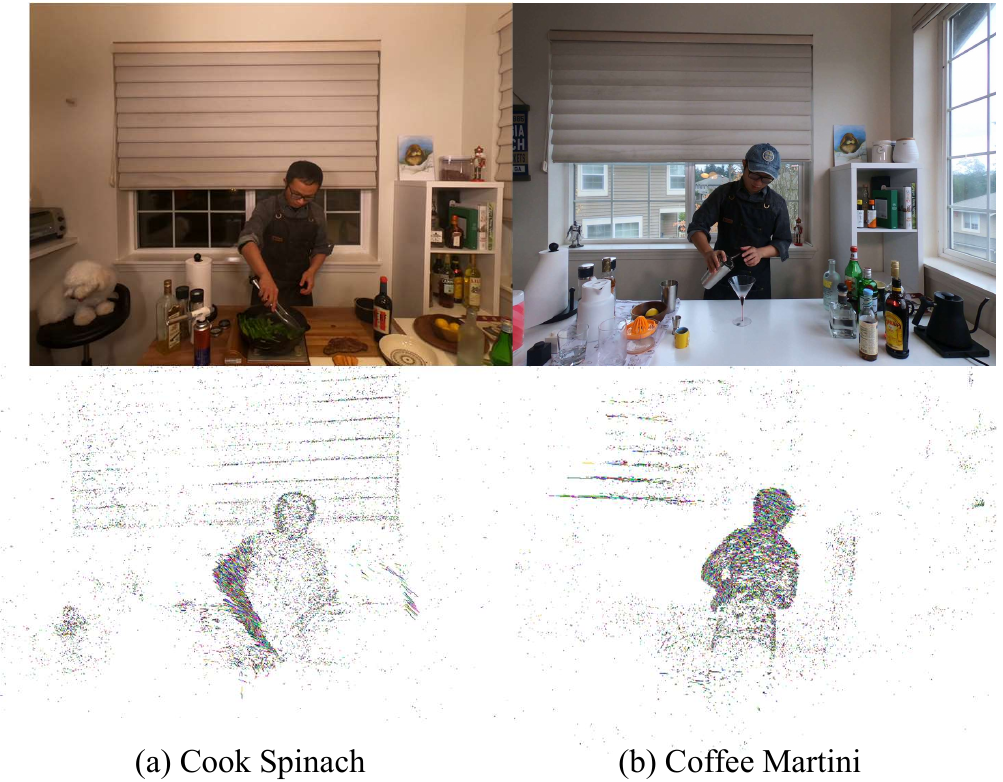}
   \caption{Visualization of tracking with 3D Gaussians. Lines in the figures of the second row stand for the trajectory of 3D Gaussians.}
      \label{fig:point_track}
\end{figure}

\paragraph{Tracking with 3D Gaussians.}Tracking in 3D is also a important task. FFDNeRF~\cite{guo2023forwardflowfornvs} also shows the results of tracking objects' motion in 3D. Different from dynamic3DGS~\cite{dynamic3dgs}, our methods even can present tracking objects in monocular settings with pretty low storage \ie 10MB in 3D Gaussians $\mathcal{G}$ and 8 MB in Gaussian deformation field network $\mathcal{F}$. Fig.~\ref{fig:point_track} shows the 3D Gaussian's deformation at certain timestamps.

\begin{figure}
   \centering
   \includegraphics[width=\linewidth]{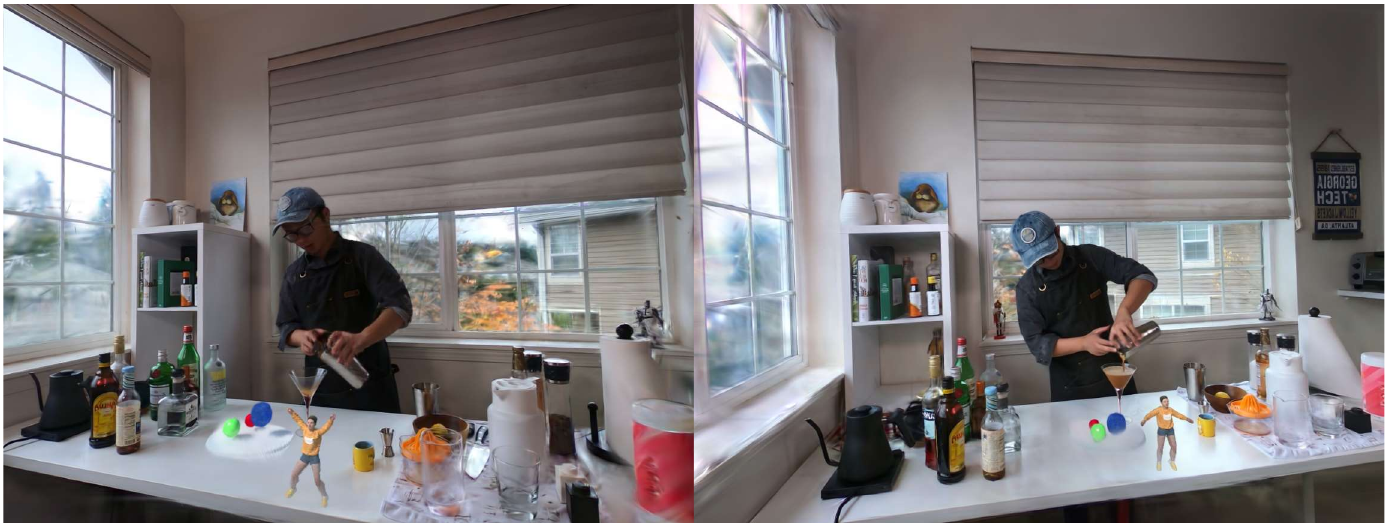}
   \caption{Visualization of composition with 4D Gaussians.}
   \label{fig: editing}
\end{figure}
\vspace{-15pt}
\paragraph{Composition with 4D Gaussians.} Similar to Dynamic3DGS~\cite{dynamic3dgs}, our proposed methods can also perform editing in 4D Gaussians, as shown in Fig.~\ref{fig: editing}. Thanks to the explicit representation of 3D Gaussians, all the trained models can predict deformed 3D Gaussians in the same space following $\mathcal{G}^\prime=\{ \mathcal{G}_1^\prime,\mathcal{G}_2^\prime,...,\mathcal{G}_n^\prime\}$ and differential rendering~\cite{yifan2019differentiablesplatting} can project all the point clouds into viewpoints by $\hat{I} = \mathcal{S}(M,\mathcal{G}^\prime)$ as referred in Sec.~\ref{subsec:4dgs}.
\begin{figure}
   \centering
   \includegraphics[width=\linewidth]{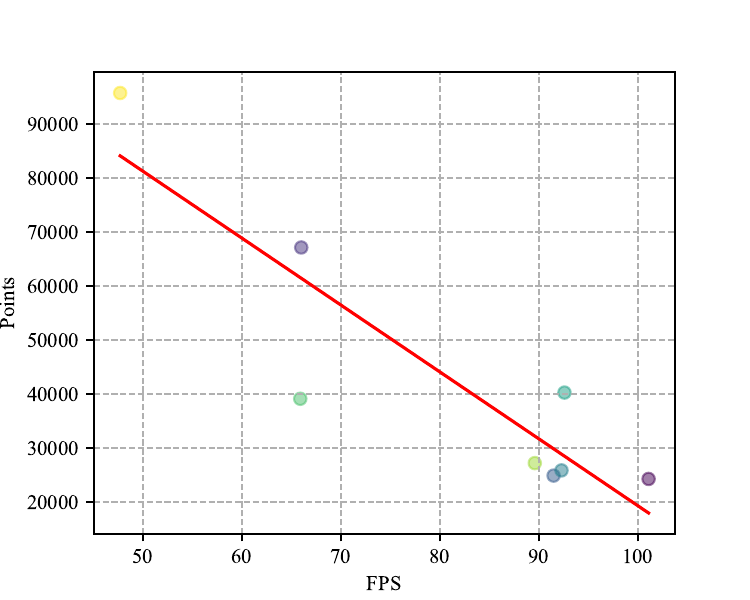}
   \caption{Visualization of the relationship between rendering speed and numbers of 3D Gaussians. All the tests are finished in the synthetic dataset.}
    \label{fig:point_fps}
\end{figure}

\paragraph{Analysis of Rendering Speed.} As shown in Fig.~\ref{fig:point_fps}, we also test the relationship between the number of 3D Gaussians and rendering speed at the resolution of 800$\times$800. We observe that if the rendered Gaussians are fewer than 30,000, the rendering speed can be up to 90 FPS on a single RTX 3090 GPU. The configuration of Gaussian deformation fields is discussed in the appendix. To achieve real-time rendering speed, we should strike a balance among all the rendering resolutions, 4D Gaussians representation including numbers of Gaussians, the capacity of the Gaussian deformation field network, and any other hardware constraints.

\subsection{Limitations}
\label{Subsec:limitation}
Though 4D-GS can indeed attain rapid convergence and yield real-time rendering outcomes in many scenarios, there are a few key challenges to address. First, large motions, the absence of background points, and the unprecise camera pose cause the struggle of optimizing 4D Gaussians. Meanwhile, it is still challenging for 4D-GS to split the joint motion of static and dynamic Gaussians under the monocular settings without any additional supervision. Finally, a more compact algorithm needs to be designed to handle urban-scale reconstruction due to the heavy querying of Gaussian deformation fields by huge numbers of 3D Gaussians. 

\section{Conclusion}
This paper proposes 4D Gaussian splatting to achieve real-time dynamic scene rendering. An efficient deformation field network is constructed to accurately model Gaussian motions and shape deformations, where adjacent Gaussians are connected via a spatial-temporal structure encoder. Connections between Gaussians lead to more complete deformed geometry, effectively avoiding avulsion. Our 4D Gaussians can not only model dynamic scenes but also have the potential for 4D objective tracking and editing.
\section*{Acknowledgments}
This work was supported by the National Natural Science Foundation of China (No. 62376102). The authors would like to thank Haotong Lin for providing the quantitative results of Im4D~\cite{lin2023im4d}.

{
    \small
    \bibliographystyle{ieeenat_fullname}
    \bibliography{main}
}

\begin{figure*}
   \centering
   \includegraphics[width=1.0\linewidth]{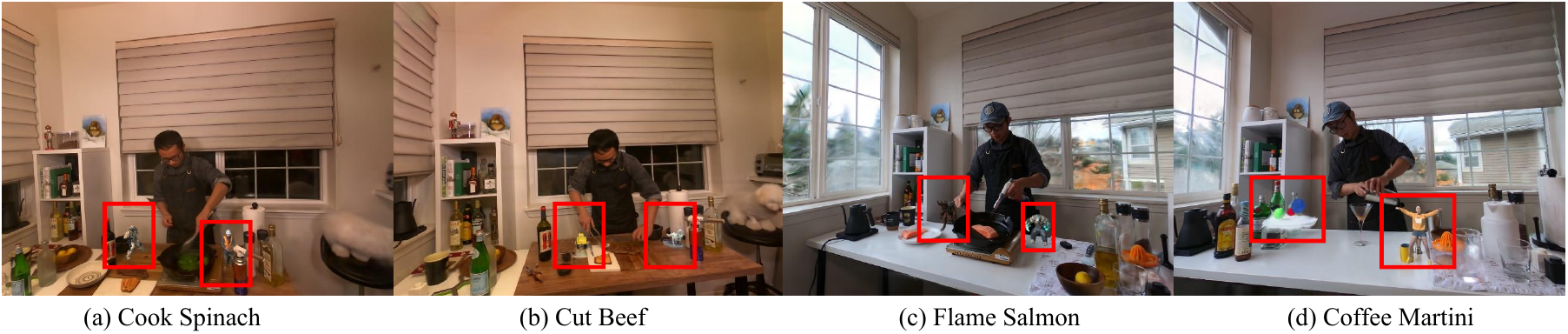}
   \caption{More visualization of composition in 4D Gaussians. (a) Composition with Punch and Standup. (b) Composition with Lego and Trex. (c) Composition with Hellwarrior and Mutant. (d) Composition with Bouncingballs and Jumpingjacks.}
    \label{fig:editing_supp}
\end{figure*}
\begin{figure}
   \centering
   \includegraphics[width=\linewidth]{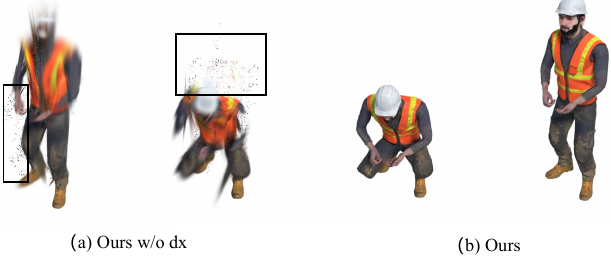}
   \caption{Visualization of ablation study about $\phi_x$.}
    \label{fig:no_dx_vis}
\end{figure}
\begin{figure}
   \centering
   \includegraphics[width=1.0\linewidth]{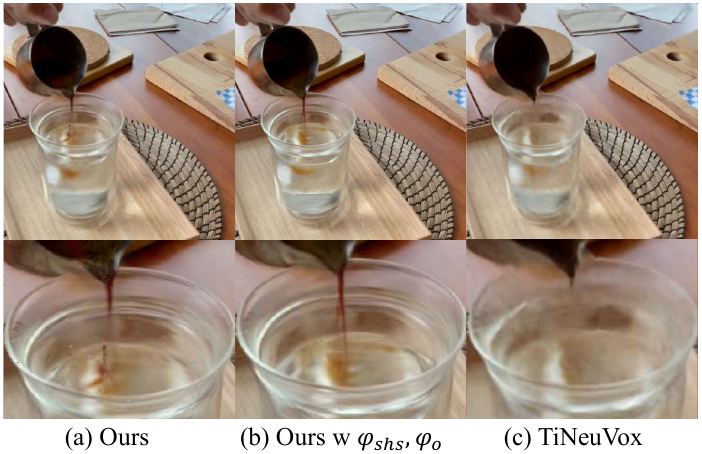}
   \caption{Visualization of ablation study in $\phi_{\mathcal{C}}$ and $\phi_{\mathcal{\alpha}}$ comparing with TiNeuVox~\cite{tineuvox}.}
    \label{fig:ab_dshs}
\end{figure}
\begin{figure*}
   \centering
   \includegraphics[width=1.0\linewidth]{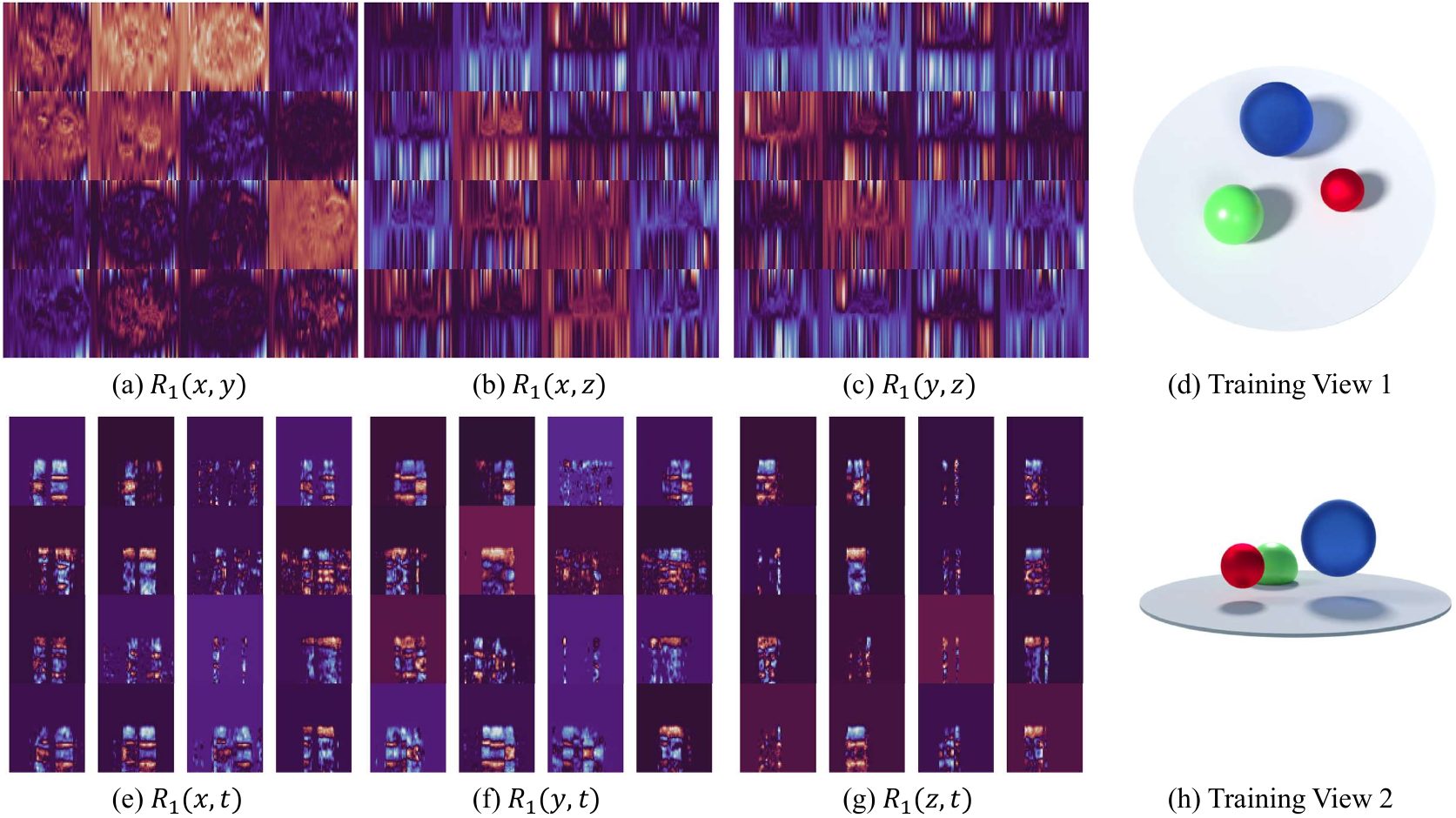}
   \caption{More visualization of the HexPlane voxel grids $R(i,j)$ in bouncing balls. (a)-(c), (e)-(f) stand for visualization of $R_1(i,j)$, where grids resolution equals to 64$\times$64.}
    \label{fig:grid_feature_vis_supp}
\end{figure*}

\begin{figure}
   \centering
   \includegraphics[width=1.0\linewidth]{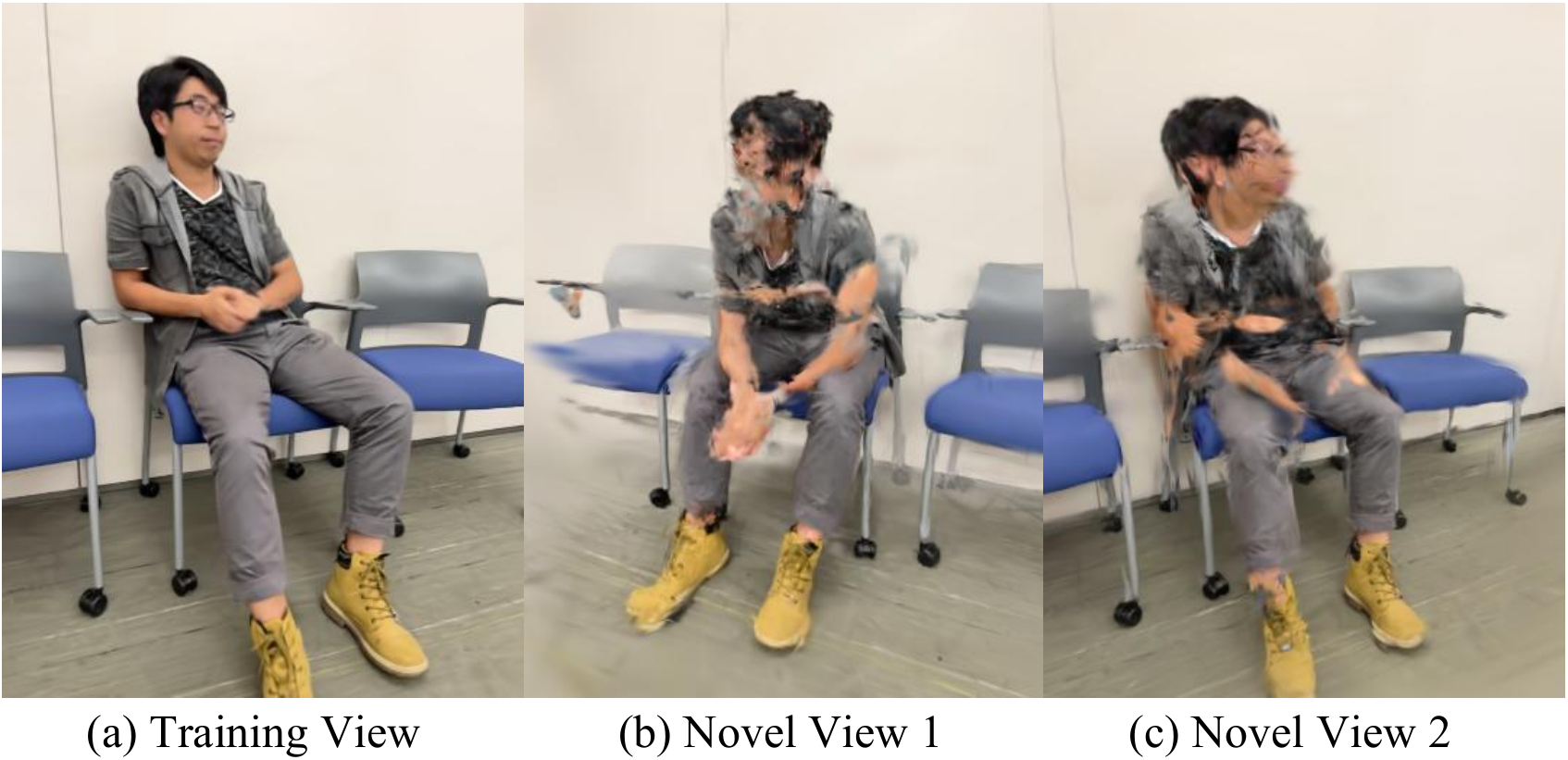}
   \caption{Novel view rendering results in the iPhone dataset~\cite{dycheck}.}
    \label{fig:dycheck_supp}
\end{figure}

\begin{figure}
   \centering
   \includegraphics[width=1.0\linewidth]{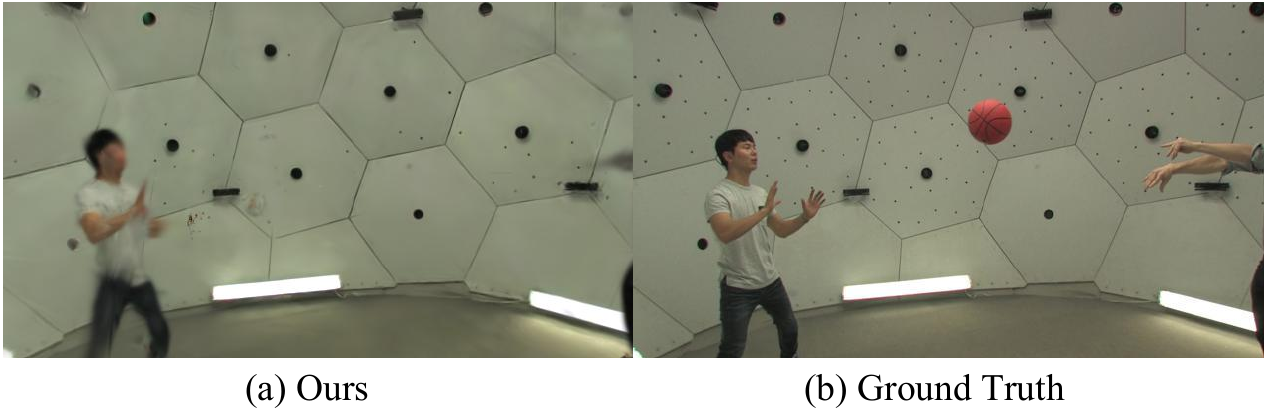}
   \caption{Rendering results on sports dataset~\cite{joo2015panopticstudio}, also used in Dynamic3DGS~\cite{dynamic3dgs}.}
    \label{fig:sports_supp}
\end{figure}

\begin{figure}
   \centering
   \includegraphics[width=1.0\linewidth]{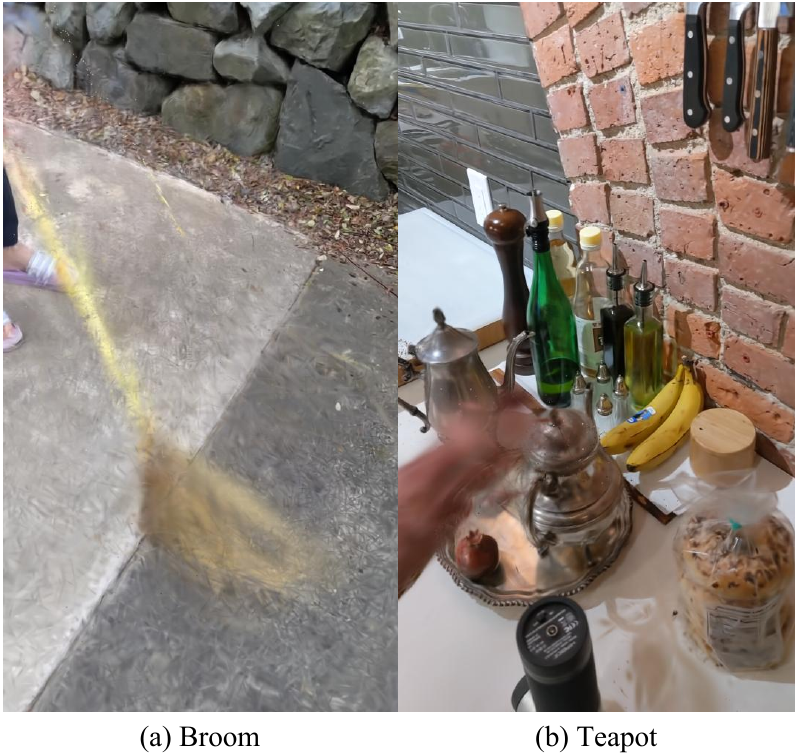}
   \caption{Failure cases of modeling large motions and dramatic scene changes. (a) The sudden motion of the broom makes optimization harder. (b) Teapots have large motion and a hand is entering/leaving the scene.}
    \label{fig:failurecases}
\end{figure}

\begin{table*}	
\centering  
\begin{threeparttable}  
\caption{Perscene results on the HyperNeRF vrig dataset~\cite{park2021hypernerf} of different models.}
\label{tab:hypernerf_comparison}
\begin{tabular}{l|cc|cc|cc|cc}  
\toprule  
\multirow{2}{*}{Method}&  
\multicolumn{2}{c}{3D Printer}&\multicolumn{2}{c}{Chicken}&\multicolumn{2}{c}{Broom}&\multicolumn{2}{c}{Banana}\cr  
\cmidrule(lr){2-3}\cmidrule(lr){4-5}\cmidrule(lr){6-7}\cmidrule(lr){8-9}
&PSNR&MS-SSIM&PSNR&MS-SSIM&PSNR&MS-SSIM&PSNR&MS-SSIM\cr  
\midrule  
Nerfies~\cite{park2021nerfies}&20.6&0.83&26.7&0.94&19.2&0.56&22.4&0.87\cr  HyperNeRF~\cite{park2021hypernerf}&20.0&0.59&26.9&0.94&19.3&0.59&23.3&0.90\cr  
TiNeuVox-B~\cite{tineuvox}&\cellcolor{pink}22.8&\cellcolor{pink}0.84&\cellcolor{yellow}28.3&\cellcolor{pink}0.95&21.5&0.69&24.4\cellcolor{yellow}&\cellcolor{yellow}0.87\cr  
FFDNeRF~\cite{guo2023forwardflowfornvs}&\cellcolor{yellow}22.8&\cellcolor{yellow}0.84&28.0&\cellcolor{yellow}0.94&21.9\cellcolor{yellow}&0.71\cellcolor{pink}&24.3&0.86\cr
3D-GS~\cite{3dgs}&18.3&0.60&19.7&0.70&20.6&0.63&20.4&0.80\cr
Ours&22.1&0.81&\cellcolor{pink}28.7&0.93&\cellcolor{pink}22.0&0.70\cellcolor{yellow}&\cellcolor{pink}28.0&\cellcolor{pink}0.94\cr
\bottomrule  
\end{tabular}  
\end{threeparttable}  
\end{table*}

\begin{table*}
\centering  
\begin{threeparttable}  
\caption{Per-scene results on the DyNeRF~\cite{li2022neural} dataset.}  
\label{tab:dynerf_comparison}  
\begin{tabular}{l|cc|cc|cc}  
    \toprule  
    \multirow{2}{*}{Method}&  
    \multicolumn{2}{c}{Cut Beef}&\multicolumn{2}{c}{Cook Spinach}&\multicolumn{2}{c}{Sear Steak}\cr  
    \cmidrule(lr){2-3}\cmidrule(lr){4-5}\cmidrule(lr){6-7}
    &PSNR&SSIM&PSNR&SSIM&PSNR&SSIM\cr  
    \midrule  
    NeRFPlayer~\cite{song2023nerfplayer}&31.83&0.928&32.06&0.930&32.31&0.940\cr
    HexPlane~\cite{hexplane}&\cellcolor{yellow}32.71&\cellcolor{pink}0.985&31.86&\cellcolor{pink}0.983&32.09&0.986\cr
    KPlanes~\cite{kplanes}&31.82&0.966&\cellcolor{pink}32.60&\cellcolor{yellow}0.966&\cellcolor{pink}32.52&\cellcolor{pink}0.974\cr
    MixVoxels~\cite{wang2023mixedvoxels}&31.30&0.965&31.65&0.965&31.43&\cellcolor{yellow}0.971\cr
    Ours&\cellcolor{pink}32.90&0.957&\cellcolor{yellow}32.46&0.949&\cellcolor{yellow}32.49&0.957\cr
\toprule  
    \multirow{2}{*}{Method}&  
    \multicolumn{2}{c}{Flame Steak}&\multicolumn{2}{c}{Flame Salmon}&\multicolumn{2}{c}{Coffee Martini}\cr  
    \cmidrule(lr){2-3}\cmidrule(lr){4-5}\cmidrule(lr){6-7}
    &PSNR&SSIM&PSNR&SSIM&PSNR&SSIM\cr  
    \midrule
    NeRFPlayer~\cite{song2023nerfplayer}&27.36&0.867&26.14&0.849&\cellcolor{pink}32.05&\cellcolor{pink}0.938\cr
    HexPlane~\cite{hexplane}&31.92&\cellcolor{pink}0.988&\cellcolor{yellow}29.26&\cellcolor{pink}0.980&-&-\cr
    KPlanes~\cite{kplanes}&32.39&0.970&\cellcolor{pink}30.44&\cellcolor{yellow}0.953&29.99\cellcolor{yellow}&\cellcolor{yellow}0.953\cr
    MixVoxels~\cite{wang2023mixedvoxels}&31.21&\cellcolor{yellow}0.970&29.92&0.945&29.36&0.946\cr
    Ours&\cellcolor{pink}32.51&0.954&29.20&0.917&27.34&0.905\cr
    \bottomrule  
\end{tabular}  
\end{threeparttable}  
\end{table*}  

\begin{table*}
\centering  
\scalebox{0.9}{

\begin{threeparttable}  
\caption{Per-scene results on synthetic datasets.}  
\label{tab:dnerf_comparison}  
\begin{tabular}{l|ccc|ccc|ccc|ccc}  
    \toprule  
    \multirow{2}{*}{Method}&  
    \multicolumn{3}{c}{Bouncing Balls}&\multicolumn{3}{c}{Hellwarrior}&\multicolumn{3}{c}{Hook}&\multicolumn{3}{c}{Jumpingjacks}\cr  
    \cmidrule(lr){2-4}\cmidrule(lr){5-7}\cmidrule(lr){8-10}\cmidrule(lr){11-13}
    &PSNR&SSIM&LPIPS&PSNR&SSIM&LPIPS&PSNR&SSIM&LPIPS&PSNR&SSIM&LPIPS\cr  
    \midrule  
    3D-GS~\cite{3dgs}&23.20&0.9591&0.0600&24.53&0.9336&0.0580\cellcolor{yellow}&21.71 & 0.8876& 0.1034& 23.20&0.9591 &0.0600\cr  
    K-Planes\cite{kplanes}&40.05&\cellcolor{yellow}0.9934&0.0322\cellcolor{yellow}&24.58&0.9520&0.0824&28.12&0.9489&0.0662&31.11&0.9708&0.0468\cr  
    HexPlane\cite{hexplane}&39.86&0.9915&0.0323&24.55&0.9443&0.0732&28.63&0.9572\cellcolor{yellow}&0.0505\cellcolor{yellow}&31.31&0.9729&0.0398\cr  
    TiNeuVox\cite{tineuvox}&\cellcolor{yellow}40.23&0.9926&0.0416&\cellcolor{yellow}27.10&\cellcolor{yellow}0.9638&0.0768&28.63\cellcolor{yellow}&0.9433&0.0636&33.49\cellcolor{yellow}&0.9771\cellcolor{yellow}&0.0408\cellcolor{yellow}\cr  		Ours&\cellcolor{pink}40.62&\cellcolor{pink}0.9942&\cellcolor{pink}0.0155&\cellcolor{pink}28.71&\cellcolor{pink}0.9733&\cellcolor{pink}0.0369&\cellcolor{pink}32.73&\cellcolor{pink}0.9760&\cellcolor{pink}0.0272&\cellcolor{pink}35.42&\cellcolor{pink}0.9857&\cellcolor{pink}0.0128\cr  
\toprule  
\multirow{2}{*}{Method}&  
    \multicolumn{3}{c}{Lego}&\multicolumn{3}{c}{Mutant}&\multicolumn{3}{c}{ Standup}&\multicolumn{3}{c}{Trex}\cr  
    \cmidrule(lr){2-4}\cmidrule(lr){5-7}\cmidrule(lr){8-10}\cmidrule(lr){11-13}
    &PSNR&SSIM&LPIPS&PSNR&SSIM&LPIPS&PSNR&SSIM&LPIPS&PSNR&SSIM&LPIPS\cr  
    \midrule  
    3D-GS~\cite{3dgs}&23.06&0.9290&0.0642&20.64&0.9297&0.0828&21.91&0.9301&0.0785&21.93&0.9539&0.0487\cr  
    K-Planes~\cite{kplanes}&\cellcolor{pink}25.49&\cellcolor{pink}0.9483&\cellcolor{pink}0.0331&32.50&0.9713&0.0362&33.10&0.9793&0.0310&30.43&0.9737&0.0343\cr  
    HexPlane~\cite{hexplane}&\cellcolor{yellow}25.10&\cellcolor{yellow}0.9388&0.0437&33.67\cellcolor{yellow}&0.980\cellcolor{yellow}2&0.0261\cellcolor{yellow}&34.40&\cellcolor{yellow}0.9839&\cellcolor{yellow}0.0204&30.67&\cellcolor{yellow}0.9749&\cellcolor{yellow}0.0273\cr  
    TiNeuVox~\cite{tineuvox}&24.65&0.9063&0.0648&30.87&0.9607&0.0474&\cellcolor{yellow}34.61&0.9797&0.0326&\cellcolor{yellow}31.25&0.9666&0.0478\cr  
    Ours&25.03&0.9376&0.0382\cellcolor{yellow}&\cellcolor{pink}37.59&\cellcolor{pink}0.9880&\cellcolor{pink}0.0167&\cellcolor{pink}38.11&\cellcolor{pink}0.9898&\cellcolor{pink}0.0074&\cellcolor{pink}34.23&\cellcolor{pink}0.9850&\cellcolor{pink}0.0131\cr  
    \bottomrule  
\end{tabular}  
\end{threeparttable}  
}
\end{table*}

\begin{table}
\centering  
\caption{Ablation Study on $\phi_{\mathcal{C}}$ and $\phi_{\mathcal{\alpha}}$, comparing with TiNeuVox~\cite{tineuvox} in Americano of 
 the HyperNeRF~\cite{park2021hypernerf} dataset.}
\begin{threeparttable}  
\label{tab:ablation_dshsdo}  
\begin{tabular}{l|cc }  
\toprule  
\multirow{2}{*}{Method}&  
\multicolumn{2}{c}{Americano}\cr  
\cmidrule(lr){2-3}
&PSNR&MS-SSIM\cr  
\midrule  
TiNeuVox-B~\cite{tineuvox}&28.4&0.96\cr  
Ours w/ $\phi_{\mathcal{C}}$,$\phi_{\mathcal{\alpha}}$ &\cellcolor{pink}31.53&\cellcolor{pink}0.97\cr
Ours&\cellcolor{yellow}30.90&\cellcolor{yellow}0.96\cr
\bottomrule  

\end{tabular}  
\end{threeparttable}  
\label{table:ab_dshsdo}

\end{table}  
\appendix

\section{Appendix}

In the supplementary material, we mainly introduce our hyperparameter settings of experiments in Sec.~\ref{Supsec:hyperparameters}. Then more ablation studies are conducted in Sec.~\ref{Supsec:ablationstudies}. Finally, we delve into the limitations of our proposed 4D-GS in Sec.~\ref{supsec:discussions}.

\subsection{Hyperparameter Settings}
\label{Supsec:hyperparameters}
Our hyperparameters mainly follow the settings of 3D-GS~\cite{3dgs}. The basic resolution of our multi-resolution HexPlane module $R(i,j)$ is set to 64, which is upsampled by 2 and 4. The learning rate is set as $1.6\times10^{-3}$, decayed to $1.6\times10^{-4}$ at the end of training. The Gaussian deformation decoder is a tiny MLP with a learning rate of $1.6\times10^{-4}$ which decreases to $1.6\times10^{-5}$. The batch size in training is set to 1. The opacity reset operation in 3D-GS~\cite{3dgs} is not used as it does not bring evident benefit in most of our tested scenes. Besides, we find that expanding the batch size will indeed contribute to rendering quality but the training cost increases accordingly.

Different datasets are constructed under different capturing settings. D-NeRF~\cite{pumarola2021dnerf} is a synthetic dataset in which each timestamp has only one single captured image following the monocular setting. This dataset has no background which is easy to train, and can reveal the upper bound of our proposed framework. We change the pruning interval to 8000 and only set a single upsampling rate of the multi-resolution HexPlane Module $R(i,j)$ as 2 because the structure information is relatively simple in this dataset. The training iteration is set to 20000 and we stop 3D Gaussians from growing at the iteration of 15000.

The Neu3D dataset~\cite{li2022neural} includes 15 -- 20 fixed camera setups, so it's easy to get the SfM~\cite{schonberger2016structurefrommotion} point in the first frame. We utilize the dense point-cloud reconstruction and downsample it lower than 100k to avoid out of memory error. Thanks to the efficient design of our 4D Gaussian splatting framework and the tiny movement of all the scenes, only 14000 iterations are needed and we can get the high rendering quality images. 

HyperNeRF dataset~\cite{park2021hypernerf} is captured with fewer than 2 cameras in feed-forward settings. We change the upsampling resolution up to $[2,4]$ and the hidden dim of the decoder to 128. Similar to other works~\cite{tineuvox,park2021hypernerf}, we found that Gaussian deformation fields always fall into the local minima that link the correlation of motion between cameras and objects even with static 3D Gaussian initialization. And we're going to reserve the splitting of the relationship in the future works. 

\subsection{More Ablation Studies}
\label{Supsec:ablationstudies}

\paragraph{Editing with 4D Gaussians.}
We provide more visualization in editing with 4D Gaussians in Fig.~\ref{fig:editing_supp}.  This work only proposes a naive approach to transformation. It is worth noting that when applying the rotation of the scenes, 3D Gaussian's rotation quaternion $r$ and scaling coefficient $s$ need to be considered. Meanwhile, some interpolation methods should be applied to enlarge or reduce 4D Gaussians.
\paragraph{Position Deformation.} We find that removing the output of the position deformation head can also model the object motion. It is mainly because leaving some 3D Gaussians in the dynamic part, keeping them small in shape, and then scaling them up at a certain timestamp can also model the dynamic part. However, this approach can only model coarse object motion and lost potential for tracking. The visualization is shown in Fig.~\ref{fig:no_dx_vis}.
\paragraph{Color and Opacity's Deformation.}
When encountered with fluid or non-rigid motion, we adopt another two output MLP decoder $\phi_{\mathcal{C}}$, $\phi_\alpha$ to compute the deformation of 3D Gaussian's color and opacity $\Delta \mathcal{C} = \phi_{\mathcal{C}}(f_d)$, $\Delta \alpha = \phi_{\mathcal{\alpha}}(f_d)$. Tab.~\ref{table:ab_dshsdo} and Fig.~\ref{fig:ab_dshs} show the results in comparison with TiNeuVox~\cite{tineuvox}. However, it is worth noting that modeling Gaussian color and opacity change may cause irrational shape changes when rendering novel views. \ie the Gaussians on the surface should move with other Gaussians but stay in the place and the color is changed, making the tracking difficult to achieve.
\paragraph{Spatial-temporal Structure Encoder.}
We explore why 4D-GS can achieve such a fast convergence speed and rendering quality. As shown in Fig.~\ref{fig:grid_feature_vis_supp}, we visualize the full features of $R_1$ in bouncingballs. It's explicit that in the $R_1(x,y)$ plane, the spatial structure of the scenes is encoded. Similarily, $R_1(x,z)$ and $R_1(y,z)$ also show different view structure features. Meanwhile, temporal voxel grids $R_1(x,t),R_1(y,t)$ and $R_1(z,t)$ also show the integrated motion of the scenes, where large motions always stand for explicit features. So, it seems that the proposed HexPlane module encodes the features of spatial and temporal information.

\subsection{More Discussions}
\label{supsec:discussions}
\paragraph{Monocular Dynamic Scene Novel View Synthesis.}
In monocular settings, the input is sparse in both camera pose and timestamp dimensions. This may cause the local minima of overfitting with training images in some complicated scenes. As shown in Fig.~\ref{fig:dycheck_supp}, though 4D-GS can render relatively high quality in the training set, the strong overfitting effects of the proposed model cause the failure of rendering novel views. To solve the problem, more priors such as depth supervision or optical flow may be needed.

\paragraph{Large Motion Modeling with Multi-Camera Settings.}
In the Neu3D~\cite{li2022neural} dataset, all the motion parts of the scene are not very large and the multi-view camera setup also provides a dense sampling of the scene. That is the reason why 4D-GS can perform a relatively high rendering quality. However, in large motion such as sports datasets~\cite{joo2015panopticstudio} used in Dynamic 3DGS~\cite{dynamic3dgs}, 4D-GS cannot fit well within short times as shown in Fig.~\ref{fig:sports_supp}. Online training~\cite{dynamic3dgs,abou2022particlenerf} or using information from other views like~\cite{lin2023im4d,wang2021ibrnet} could be a better approach to solve the problem with multi-camera input.

\paragraph{Large Motion Modeling with Monocular Settings.}
4D-GS uses a deformation field network to model the motion of 3D Gaussians, which may fail in modeling large motions or dramatic scene changes. This phenomenon is also observed in previous NeRF-based methods~\cite{tineuvox,pumarola2021dnerf,park2021hypernerf,li2022neural}, producing blurring results. Fig.~\ref{fig:failurecases} shows some failed samples. Exploring more useful priors could be a promising future direction.


%


\end{document}

%% file: preamble.tex
\usepackage[dvipsnames]{xcolor}
